%% file: eccv_2022_aleks.tex
\documentclass[runningheads]{llncs}
\usepackage{graphicx}
\usepackage{svg}

\usepackage{tikz}
\usepackage{comment}
\usepackage{amsmath,amssymb} 
\usepackage{color}
\usepackage{xspace}

\usepackage[accsupp]{axessibility}  

\usepackage{adjustbox}
\usepackage{moresize}
\usepackage{stfloats}
\usepackage{multirow}
\usepackage{amssymb}
\usepackage{pifont}
\usepackage{booktabs}
\definecolor{citecolor}{RGB}{34,139,34}

\usepackage[pagebackref=true,breaklinks=true,colorlinks,citecolor=citecolor,bookmarks=false]{hyperref}
\usepackage{subcaption}

\input{macros}

\begin{document}
\pagestyle{headings}
\mainmatter
\def\ECCVSubNumber{514}  

\title{\papertitle}


\titlerunning{\methodabr}
%
\author{Aleksandr Kim \and
Guillem Brasó \and
Aljoša Ošep \and Laura Leal-Taix{\'e}}
\authorrunning{A. Kim et al.}
%

\institute{Technical University of Munich, Germany \\
\email{\{aleksandr.kim,guillem.braso,aljosa.osep,leal.taixe\}@tum.de}}
\maketitle

\begin{abstract}
\begin{itemize}

Most (3D) multi-object tracking methods rely on appearance-based cues for data association.  
By contrast, we investigate \textit{how far we can get} by only encoding geometric relationships between objects in 3D space as cues for data-driven data association.
We encode 3D detections as nodes in a graph, where spatial and temporal pairwise relations among objects are encoded via \textit{localized polar} coordinates on graph edges. 
This representation makes our geometric relations invariant to global transformations and smooth trajectory changes, especially under non-holonomic motion. This allows our graph neural network to learn to effectively encode \textit{temporal and spatial} interactions and fully leverage contextual and motion cues to obtain final scene interpretation by posing data association as edge classification. 
We establish a new state-of-the-art on nuScenes dataset and, more importantly, show that our method, \methodabr, generalizes remarkably well across different locations (Boston, Singapore, Karlsruhe) and datasets (nuScenes and KITTI). 
%
\end{itemize}
\end{abstract}

%
\section{Introduction}

Intelligent agents such as autonomous vehicles need to understand dynamic objects in their surroundings to safely navigate the world. 
3D multi-object tracking (MOT) is, therefore, an essential component of autonomous intelligent systems.

State-of-the-art methods leverage the representational power of neural networks to learn appearance models~\cite{weng20CVPR,zeng21iros} or regress velocity vectors~\cite{yin2021center} as cues for data association. 
While powerful, such methods need to be trained for the specific environments in which they are deployed. 
Methods such as~\cite{Weng20iros,chiu2020probabilistic} rely on motion as the key cue for association and can thus generalize across different environments. However, performance-wise they lag behind data-driven methods as they treat individual objects in isolation and do not consider their interactions.

In this work, we investigate \textit{how far we can get} by learning to track objects given \textit{only geometric cues} in the form relative pose differences between 3D bounding boxes \textit{without} relying on any appearance information. 
This approach is not coupled to any specific object detector, sensor modality, or region-specific appearance models. Consequentially, it generalizes well across different environments and geographic locations, as we experimentally demonstrate. 
By contrast to prior work~\cite{Weng20iros} that rely on individual object motion as the main association cue, our method can learn how objects move as a group, taking into account their interactions to better adapt to dynamic environments and crowded scenarios.

\begin{figure}[t]
\centering
\includegraphics[width=1\linewidth]{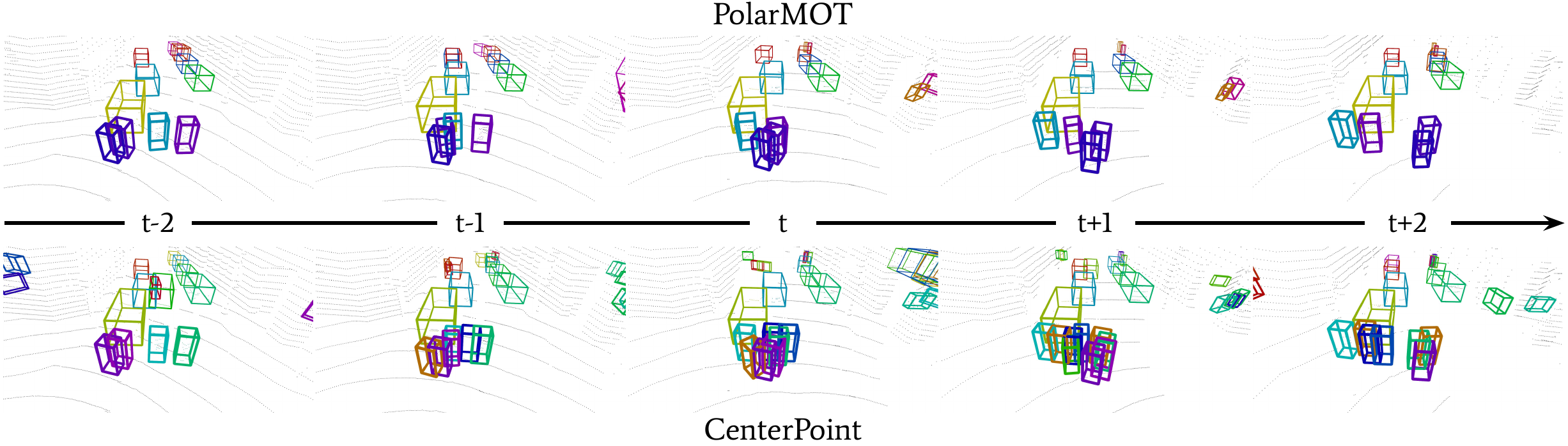}
\caption{\methodabr tracks objects in 3D (offline and online) using a graph neural network that learns to associate 3D bounding boxes over time solely based on their relative geometric features and spatio-temporal relationships. Considering such object interactions improves \methodabr's ability to handle scenarios with high occlusions (pedestrian crossing) compared to others, \eg CenterPoint.}
\vspace{-20pt}
\label{fig:teaser_qualitative}
\end{figure}

\begin{figure}[t]
\centering
\includegraphics[width=1\linewidth]{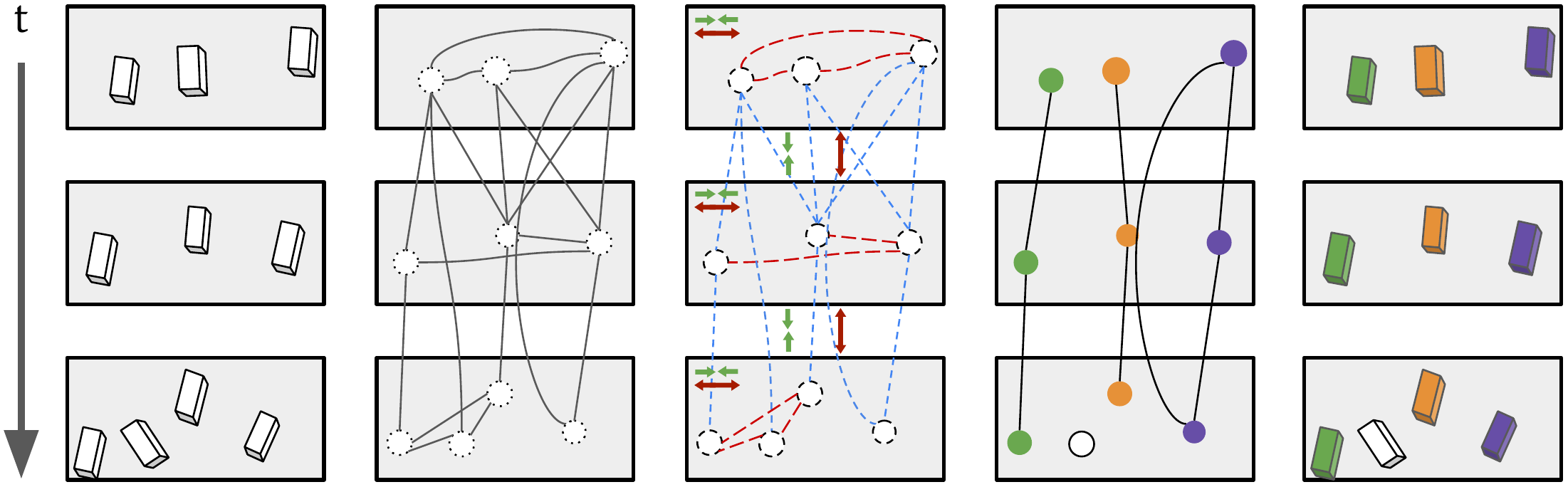}
\caption{Given a set of 3D bounding boxes in a sequence, \methodabr constructs a graph encoding detections as nodes, and their geometric relations as \textcolor{darkred}{\textit{spatial}} and \textcolor{lightblue}{\textit{temporal}} edges. After refining edge features via message passing with wider spatial and temporal context, we classify edges to obtain object track predictions.}
\vspace{-20pt}
\label{fig:teaser_pipeline}
\end{figure}

As graphs are a natural representation for such long-term agent interactions, we represent our scene as a \textit{sparse}, \textit{multiplex} graph, which is then processed by a graph neural network (Fig. \ref{fig:teaser_pipeline}). 
We encode 3D detection as nodes, while edges represent their \textit{spatial} and \textit{temporal} relations and denote possible associations and influences.  
After several message passing steps in this graph, our neural network outputs binary classifications for all \textit{temporal} edges. All nodes connected with a positive edge form a track and thus get assigned a consistent track ID~\cite{Braso20CVPR}. 


One of our main insights is that encoding pairwise geometric relations, \ie, edge features, in \textit{localized polar} instead of Cartesian space is the key to good generalization and robustness and enables our model to effectively learn to track by \textit{solely} relying on geometric cues.
The pair-specific nature of the features makes them invariant to global transforms. Thus, features encoding relations between different detections of an object that moves consistently will be \textit{stable}, regardless of the route geometry or point of reference. 
Such polar representation also naturally encodes non-holonomic motion prior, \ie, a motion that is constrained by heading direction and is well parametrized by heading angle and velocity. 

%
%

We evaluate our method, \methodabr, on KITTI~\cite{Geiger12CVPR} and nuScenes datasets~\cite{Caesar20CVPR}. Our ablations reveal that our proposed graph structure and edge feature parametrization are pivotal for our final model achieving {$66.4$} average AMOTA, setting a new state-of-the-art on nuScenes MOT dataset among lidar-based methods. More importantly, we show that our learned tracker, which relies only on geometry leads to strong generalization across different geographic regions (Boston, Singapore, Karlsruhe) and datasets (nuScenes, KITTI), without fine-tuning. 
We are not suggesting to \textit{not use} appearance cues, but point out that we can get very far with a minimalistic, graph-based tracking, solely relying on geometric cues and inferring object properties implicitly from interactions.

To \textbf{summarize}, (i) we propose a minimalistic, graph-based 3D tracker that relies only on geometric cues and, without any bells and whistles or image/lidar input, establish new state-of-the-art on the nuScenes dataset; 
(ii) we suggest a graph-based representation of the scene that encodes temporal and spatial relations via a localized polar representation. This is not only to achieve state-of-the-art performance on benchmarks but, more importantly, is the key to strong generalization; 
(iii) as efficiency and online operation are crucial for robotic/AV scenarios, we construct a sparse graph and only establish links that are feasible based on maximal possible velocity and show how such sparse graphs can be constructed in an online fashion.
We hope our code and models, available at \url{\githuburl}, will run on friendly future robots!

%

\section{Related Work}

This section reviews relevant related work in 2D and 3D multi-object tracking (MOT) based on the well-established \textit{tracking-by-detection} paradigm. 

\PAR{2D MOT.} Early methods for vision-based MOT rely on hand-crafted appearance models~\cite{Milan14TPAMI,Choi15ICCV} and motion cues~\cite{Milan14TPAMI,leal14cvpr}, in conjunction with optimization frameworks for data association that go beyond simple bi-partite matching~\cite{Kuhn55NRLQ}. 
These include quadratic pseudo-boolean optimization~\cite{Leibe08TPAMI}, graph-based methods~\cite{Pirsiavash11CVPR,Schulter17CVPR,Zhang08CVPR,ButtCollins13CVPR,Brendel11CVPR}, conditional random fields~\cite{Milan14TPAMI,Choi15ICCV} and lifted multi-cuts~\cite{Tang15CVPR}. 
Orthogonally, the community has also been investigating lifting data association from image projective space to 3D via depth maps~\cite{Leibe08TPAMI,Leibe08IJCV,Osep17ICRA,Osep18ICRA,luiten20RAL,Sharma18ICRA}. 

In the era of deep learning, the community has been focusing on learning strong appearance models~\cite{LealTaixe16CVPRW,Voigtlaender19CVPR,Son17CVPR}, future target locations~\cite{Bergmann19ICCV,xu20cvpr} or predicting offset vectors as cues for association~\cite{zhou2020tracking}. 
Recently, we have witnessed a resurgence in graph-based approaches for tracking. MPNTrack~\cite{Braso20CVPR} encodes 2D object detections as nodes while edges represent possible associations among them. Then, a message passing~\cite{gilmer2017neural} graph neural network (GNN)~\cite{Gori05IJCNN} is used to update node/edge representation with temporal context, as needed for reliable binary edge classification to obtain final tracks.  
Just as \cite{Braso20CVPR}, we (i) encode detections as nodes, while edges represent hypothetical associations, and we (ii) learn to classify edges to obtain the final set of tracks. Different to \cite{Braso20CVPR}, we tackle 3D MOT using \textit{only} geometric cues. Our model encodes both, temporal and spatial relations and, importantly, encodes them via \textit{localized polar coordinates}, that encode non-holonomic motion prior. Moreover, beyond \textit{offline} use-case, we show how to construct \textit{sparse} graphs in \textit{online} fashion, as needed in mobile robotics. 


\PAR{3D MOT.} 
Early methods for 3D detection and tracking~\cite{dellaert1997robust,Petrovskaya09AR} model vehicles as cuboids or rely on bottom-up point cloud segmentation~\cite{Teichman11ICRA,Moosmann13ICRA,Held14RSS} and track objects using (extended) Kalman filter. 
Thanks to developments in deep point-based representation learning \cite{Qi17CVPR_pointnet,Qi17NIPS,Thomas19ICCV,zhou2018voxelnet}, we nowadays have strong backbones, as needed for 3D object detection \cite{Qi17CVPR_frustum,Shi19CVPR,zhou2018voxelnet,yan2018second,Lang19CVPR,yin2021center}, tracking~\cite{Frossard18ICRA,Weng20iros,chiu2020probabilistic} and segmentation~\cite{aygun21cvpr}. 
AB3DMOT~\cite{Weng20iros} shows that well-localized lidar-based 3D object detections work very well in conjunction with a simple Kalman filter based tracking framework~\cite{bewley16icip}. Association can be performed based on 3D bounding box overlap or centroid-based Mahalanobis distance~\cite{chiu2020probabilistic} to increase robustness to lower frame rates. 
State-of-the-art CenterPoint \cite{yin2021center} learns to detect objects as points and regresses velocity vectors needed for the association.
Similar to ours, OGR3MOT\cite{zaech2022learnable} also tackles 3D MOT using message passing networks based framework, proposed by~\cite{Braso20CVPR}. However, it represents detections \textit{and} tracks as two distinct types of nodes, thus effectively maintaining two subgraphs. 
\textit{Different} to that, \textit{our} \methodabr implicitly derives node embeddings from relational (edge) features parametrized via proposed local polar coordinates. 
A recent body of work proposes to fuse image and lidar data~\cite{Kim21ICRA,zeng21iros,weng20CVPR}. 
GNN3DMOT~\cite{weng20CVPR} utilizes GNNs to learn appearance and motion features for data association jointly. This approach relies on Hungarian algorithm to perform data association based on the learned features. 
\textit{Different} to that, we \textit{only} use 3D geometric cues and perform association directly via edge classification in end-to-end manner.

\section{Message Passing Networks for Multi-Object Tracking}
\label{sec:preliminary_mpntrack}

Our work is inspired by MPNTrack \cite{Braso20CVPR}, an image-based MOT method that we summarize for completeness. MPNTrack models detections as graph nodes and represents possible associations via edges. After propagating features via neural message passing \cite{gilmer2017neural}, edges are classified as active/inactive.


MPNTrack processes a clip of frames with detected objects, and outputs classifications of links between them. 
For each clip, input is a set of object detections $\mathcal{O} = \{o_i\}_{i=1}^n$, represented via an (appearance) embedding vector, 2D position and timestamp. 
Then, a graph $G=(V, E)$ where $V= \mathcal{O}$ and $E \subset \mathcal{O} \times \mathcal{O}$.  
%
%
MPNTrack encodes only inter-frame edges and heuristically prunes them for sparsity. 
%
Each node $o_i\in V$ and edge $e_{ij}\in E$ (connecting $o_i$, $o_j$) have corresponding initial embeddings $h_i^{(0)}$ and $h_{(i,j)}^{(0)}$ obtained from appearance and position cues. 
These embeddings are propagated across the graph via neural message passing for a fixed number of iterations $L$
to obtain updated edge features $h_{(i,j)}^{(1,\dots,L)}$. 

More specifically, edge embeddings are updated at each message passing step based on embeddings from the previous step of the edge itself and its neighboring nodes. 
Nodes are then updated based on previous embeddings of the node and neighboring edges, which are aggregated separately for forward and backward directions in time.
After $L$ message passing updates, edges are classified based on their embeddings from all steps, $h_{(i,j)}^{(1, \dots, L)}$, where positive classification implies that two connected detections are part of the same track (identity). 
%

%

\begin{figure}[t]
\centering
    \begin{subfigure}[t]{0.32\textwidth}
        \includegraphics[width=\linewidth]{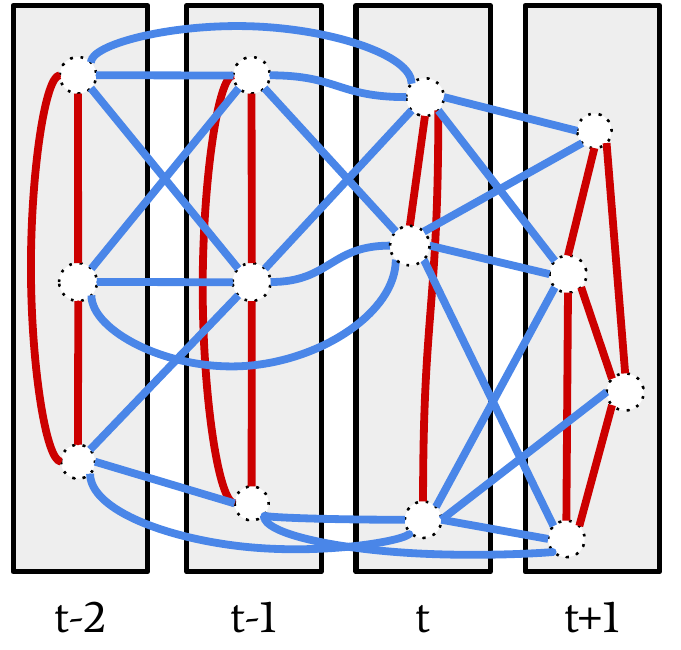}
        \caption{Full (sparse) graph.}
        \label{subfig:overview_structure}
    \end{subfigure}
    \begin{subfigure}[t]{0.32\textwidth}
        \includegraphics[width=\linewidth]{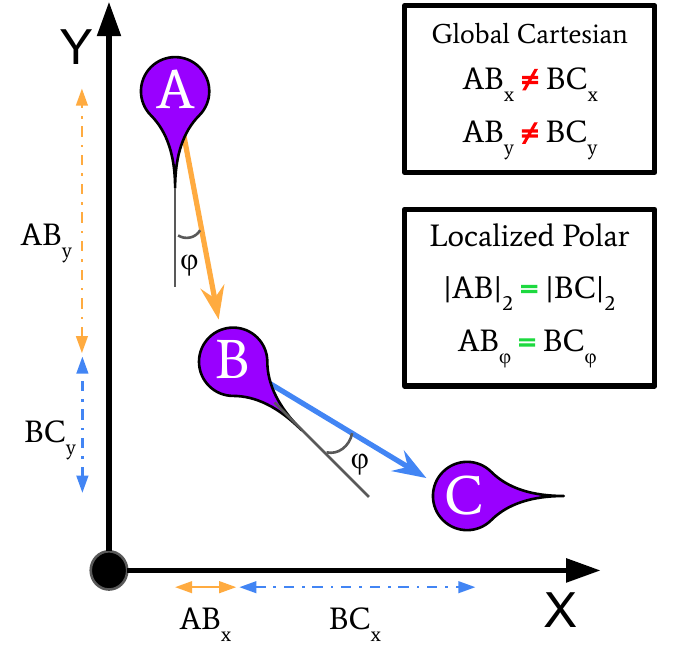}
        \caption{Relational features.}
        \label{subfig:overview_polar}
    \end{subfigure}
    \begin{subfigure}[t]{0.32\textwidth}
        \includegraphics[width=\linewidth]{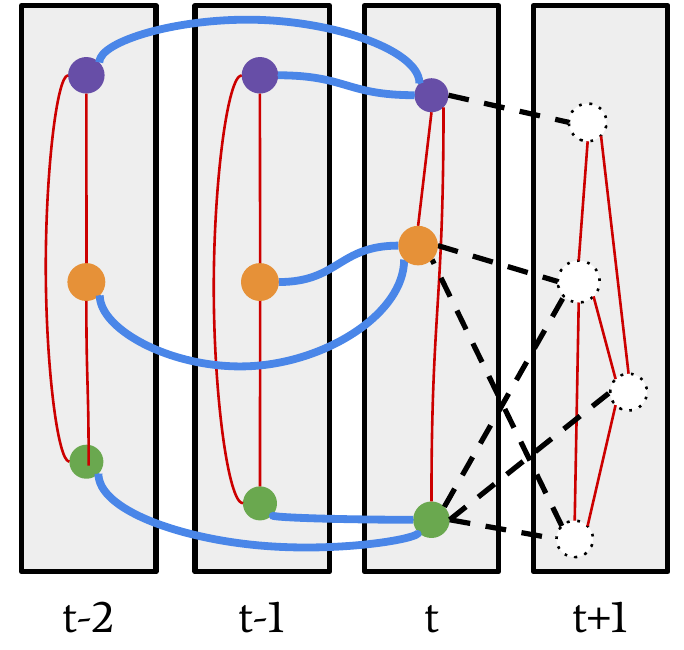}
        \caption{Online construction.}
        \label{subfig:overview_online}
    \end{subfigure}
\caption{
    The key contributions of our work: multiplex input graph with \textcolor{lightblue}{\textit{inter-frame (temporal)}} and \textcolor{darkred}{\textit{intra-frame (spatial)}} edges, relative geometric features in localized polar coordinates and continuously evolving online graph construction.
}
\vspace{-20pt}
\label{fig:overview_key_contribution}
\end{figure}

\section{PolarMOT}
\label{sec:method}
In this section, we provide a high-level overview of our \methodabr, followed by a detailed discussion of our key ideas and components. 

\subsection{Method Overview}

\methodabr encodes the overall geometric configuration of detected objects and the relative pose changes between them as primary cues for tracking. 

\PAR{Sparse graph-based scene representation.} 
We use a graph to represent object detections as nodes, and their geometric relations as edges. We encode both \textit{temporal} and \textit{spatial} relations among objects in two levels of our \textit{sparse} \textit{multiplex} graph (Fig.~\ref{subfig:overview_structure}). 
The \textcolor{lightblue}{\textit{temporal}} level connects nodes across different frames with \textcolor{lightblue}{inter-frame edges}, while the \textcolor{darkred}{\textit{spatial}} level connects nodes of the same frame with \textcolor{darkred}{intra-frame edges}. 
For sparsity, we \textit{only} link nodes that are mutually reachable based on the maximal velocity of their semantic class.

\PAR{\relativefeaturescapital}
We encode geometric relations among nodes via a localized polar-based representation (Fig.~\ref{subfig:overview_polar}). 
Our parametrization is not based on a shared global coordinate frame but is specific to each pair's local frame. This makes our parametrization \textit{invariant} to the reference world frame, and induces a \textit{non-holonomic} (directional) motion prior.

\PAR{Learning representation via message passing.} 
We follow a message passing procedure \cite{gilmer2017neural} to iteratively update our features through alternating edge/node updates (Fig.~\ref{fig:teaser_pipeline}, \textit{center}). Node features are not extracted from input detections. Instead, they are learned \textit{implicitly} through mutual object interactions. 

\PAR{Edge classification.}
To obtain object tracks, we classify our \textcolor{lightblue}{\textit{temporal}} edges using a multi-layer perceptron based on final edge representations (Fig.~\ref{fig:teaser_pipeline}, \textit{right}).

\PAR{Online graph construction.} 
For \textit{online} tracking, we continuously evolve our input graph after each frame, while maintaining both sparsity and high connectivity (Fig.~\ref{subfig:overview_online}). 
For online processing, we connect past track predictions directly to the most recent detections, thus allowing our network to infer historical context effectively via message passing on the graph topology. 

\subsection{Message Passing on a Sparse Multiplex Graph}
\label{subsec:graph_scene_representation}

\PAR{Representation.} 
Following~\cite{Braso20CVPR}, we represent each individual object detection as a \textit{node} of the graph with \textit{edges} encoding relations between objects. More precisely, we model \textcolor{lightblue}{\textit{temporal}} and \textcolor{darkred}{\textit{spatial}} relations between objects (graph nodes) by encoding our 3D detections via a  \textit{sparse}, undirected \textit{multiplex} graph with two distinct levels / edge types (Fig.~\ref{subfig:overview_structure}). 
In the \textcolor{lightblue}{\textit{first level}} we model temporal connections via undirected \textcolor{lightblue}{\textit{inter-frame}} edges between nodes across frames \cite{Braso20CVPR}, which describe potential motion of objects. 
Each edge connects two distinct detections across frames. The edge features denote the likelihood of an object moving from pose/node A to pose/node B in the elapsed time. %
In the \textcolor{darkred}{\textit{second level}}, we introduce undirected \textcolor{darkred}{\textit{intra-frame}} edges (\ie, links between nearby detections from the same frame) to model spatial context. 
These edges act as pathways to exchange frame-specific context and are meant to express the mutual influence of moving targets' motion patterns. This should intuitively help with difficult and ambiguous cases that arise in crowded scenarios. 
Both inter- and intra-frame edges connect the same set of nodes, but convey semantically different information, effectively turning our graph into a multiplex network with two distinct levels. 
\textcolor{darkred}{\textit{Intra-frame}} edges are excluded from edge classification.

\PAR{Sparse graph construction.}
To handle occlusions, we connect nodes across any number of frames. Doing this naively would potentially result in a prohibitively dense graph. 
To ensure graph sparsity, we take full advantage of the 3D domain, and rely on physical constraints to establish only \textit{relevant}, physically-plausible spatial or temporal relations.
For \textcolor{lightblue}{\textit{inter-frame edges}}, if the physical distance between two detections is greater than the maximal velocity of the detected class, we consider it impossible for them to belong to the same object and do not form an edge.
For \textcolor{darkred}{\textit{intra-frame edges}}, we allow distance up to twice the maximal velocity to connect objects that could collide in the next frame if moving towards each other and, therefore, influence each other's movement. 

%
%
%

%

\PAR{Message passing.} 
Following the general message passing framework, our model performs a fixed number of alternating \textit{edge} and \textit{node} updates {to obtain enhanced node/edge representation, as needed for edge classification}.  

\noindent \textit{Edge update:} At every message passing iteration $l=1,\dots, L$, we first update {edge embeddings} by learning to fuse the edge and connected node embeddings from the previous message passing iteration. 
More precisely, we update edge embedding $h_{(i,j)}$ between nodes $o_i$ and $o_j$ as follows: 
\begin{align}
\scriptsize
h_{(i, j)}^{(l)} =  \mlp_{\text{edge}}\left( \left[ h_{i}^{(l-1)}, h_{(i, j)}^{(l-1)}, h_{j}^{(l-1)} \right] \right).
\normalsize
\label{eq:edge_update}
\end{align}
%
%
%
During each update, the edge embedding from the previous message passing iteration $l-1$, \ie, $h_{(i, j)}^{(l-1)}$, is concatenated with embeddings $(h_{i}^{(l-1)}$, $h_{j}^{(l-1)})$ of the connected nodes.
The result is then fused by $\mlp_{\text{edge}}$, a Multi-Layer Perceptron (MLP), which produces the fused edge features $h_{(i, j)}^{(l)}$ (Eq. \ref{eq:edge_update}). 
This edge embedding update is identical for all edge types and directions.

%
\noindent \textit{Node update:}
To update nodes, first, we construct messages for each edge $h_{(i, j)}^{(l)}$ via learned fusion of \textit{its updated edge} and \textit{neighboring nodes'} features:
\begin{align}
\scriptsize
            m_{(i, j)}^{(l)} =\begin{cases} 
            \textcolor{lightblue}{\mlp_{\text{past}}}\left([h_i^{(l-1)}, h_{(i, j)}^{(l)}, h_{j}^{(l-1)}]\right) \text{if } t_j<t_i,\\[7pt]
            \textcolor{darkred}{\mlp_{\text{pres}}}\left([h_i^{(l-1)}, h_{(i, j)}^{(l)}, h_{(j)}^{(l-1)}]\right) \text{ if } t_i=t_j, \\[7pt]
            \textcolor{lightblue}{\mlp_{\text{fut}}} \text{ }\left([h_i^{(l-1)}, h_{(i, j)}^{(l)}, h_{j}^{(l-1)}]\right) \text{   if } t_j>t_i.  
          \end{cases}\label{eq:node_update_1}
\normalsize
\end{align}
Via \textcolor{lightblue}{\textit{inter-frame edges}} we produce two messages for each temporal direction: we produce \textit{past} features via \textcolor{lightblue}{$\mlp_{past}$} and \textit{future} features via \textcolor{lightblue}{$\mlp_{fut}$} 
(only in offline tracking settings).
At the same time, we produce \textit{present} features via \textcolor{darkred}{$\mlp_{pres}$} for each \textcolor{darkred}{\textit{intra-frame}}, spatial, edge.
Using separate MLPs allows our model to learn different embeddings for each type of relationship. 

\noindent \textit{Node aggregation:} Then, each node $h_i^{(l)}$ \textit{aggregates} incoming messages using an element-wise max operator, separately for past, present and future edges to maintain contextual awareness. The resulting aggregated vectors are concatenated and fused via $\mlp_{node}$:
\begin{align}
\scriptsize
h_i^{(l)} =  \mlp_{\text{node}} (  [  \textcolor{lightblue}{\max_{t_j<t_i} m^{(l)}_{(i, j)}}, \textcolor{darkred}{\max_{t_i=t_j} m^{(l)}_{(i, j)}}, \textcolor{lightblue}{\max_{t_j>t_i} m^{(l)}_{(i, j)}} ]  ). 
\label{eq:node_update_2}
\normalsize
\end{align}
As we do not use any object-specific information (\eg appearance), we rely on the model to implicitly learn node features from their interactions.

\noindent \textit{Initialization:} \textit{We initialize} edge embeddings with relational features (as detailed in Sec.~\ref{sec:polar_encoding}) processed by $\mlp_{edge\_init}$ (Eq.~\ref{eq:edge_initial}). These are then directly aggregated to produce \textit{initial node embeddings} via $\mlp_{node\_init}$ (Eq. \ref{eq:node_initial}):
\begin{align}
\scriptsize
&
\scriptsize
(\text{initial edge}) &
\scriptsize
h_{(i, j)}^{(1)} 
\scriptsize
&  
\scriptsize
=
\mlp_{\text{edge\_init}}(h_{(i, j)}^{(0)}),   \label{eq:edge_initial}\\
&
\scriptsize
(\text{initial node}) &
\scriptsize
h_i^{(1)} &
\scriptsize
=  \mlp_{\text{node\_init}} (  [  \textcolor{lightblue}{\max_{t_j<t_i} h^{(1)}_{(i, j)}}, \textcolor{darkred}{\max_{t_i=t_j} h^{(1)}_{(i, j)}}, \textcolor{lightblue}{\max_{t_j>t_i} h^{(1)}_{(i, j)} } ]  ).  
\label{eq:node_initial} 
\normalsize
\end{align}

\PAR{Tracking via edge classification.} 
After graph construction and message passing, we classify \textcolor{lightblue}{\textit{inter-frame}} edges, which encode \textit{temporal} relations. Positive classification implies the same object identity for both detections (\ie, a consistent track ID).
Any edge connecting nodes of the same track should be labeled as positive, regardless of the time difference between the connected nodes.

\subsection{Localized Relational Polar Encoding}
\label{sec:polar_encoding}

Depth sensors do not provide comparatively rich appearance information as images; however, they do provide accurate distance measurements to objects. Therefore, we fully focus on efficiently representing relative poses via edge features and learn node features implicitly. This leads to the question of how to represent geometric relations between objects efficiently? We start by representing each 3D detection as an oriented point with two planar coordinates $x, y$ and orientation $\phi$ around vertical axis $z$, as differences in elevation, tilt, and spatial dimensions are insignificant between objects of the same class.

\PAR{Global Cartesian coordinates.} 
Traditionally, spatial geometric relations between objects are parametrized in Cartesian coordinates relative to some reference frame, encoding an edge $h^{(0)}_{i,j}$ connecting nodes $o_i$ and $o_j$ as: 
\begin{equation}
\scriptsize
\label{eq:cartesian}
    {
    h^{(0)}_{i,j} = \Delta (o_i, o_j) 
    = 
    \begin{bmatrix}
    \Delta x \\
    \Delta y \\
    \dots
    \end{bmatrix}
    = 
    \begin{bmatrix}
    o_i^x - o_j^x \\
    o_i^y - o_j^y \\
    \dots
    \end{bmatrix}
    }.
\normalsize
\end{equation}
As shown in Fig.~\ref{subfig:overview_polar}, for \textit{identical} pose changes $\overrightarrow{AB}$ and $\overrightarrow{BC}$, global Cartesian features are \textit{not identical}, as they depend on the orientation of the reference frame. \textit{Identical motion non-intuitively leads to different relational features.}
\PAR{Localized polar coordinates.} 
We propose a different representation, which depends (i) only on the relevant pair of objects, $o_i$ and $o_j$, and (ii) is more suitable for encoding directional non-holonomic motion. 
Our parametrization (Fig.~\ref{subfig:overview_polar}) expresses differences in poses $A$ and $B$ through a velocity vector, expressed in \textit{polar coordinates}, where the center of the first detection $A$ is the \textit{pole} (\ie, the origin of the polar coordinate frame), and its heading direction (downward) is the \textit{polar axis}. 
This vector includes two components: \textit{velocity} $|AB|$ (\ie, distance between objects by detection time difference), and \textit{polar angle} $\varphi$, \ie, the angle between $\overrightarrow{AB}$ and the polar axis of $A$ (downward vector). 
We also include differences in object orientation ($o^\phi$) and detection time difference ($o^t$):
\begin{equation}
\scriptsize
\label{eq:polar}
    h^{(0)}_{i,j} 
    = \Delta (o_i, o_j) 
    = 
    \begin{bmatrix}
    v \\
    \varphi_{i, j} \\
    \Delta \phi\\
    \Delta t 
    \end{bmatrix}
    = 
    \begin{bmatrix}
    \frac{\Vert o_i - o_j \Vert_2}{\Delta t} \\
    \angle \left({\vec{o}_i^{~head}, ~\vec{o_j} - \vec{o_i}} \right ) \\
    o_i^\phi - o_j^\phi \\
    o_i^t - o_j^t
    \end{bmatrix}.
\normalsize
\end{equation}
%
We remove the dependency on an arbitrary reference frame by computing each feature relative to individual localized frames. 
\textit{Now, identical motion between $\overrightarrow{AB}$ and $\overrightarrow{BC}$ leads to identical feature representation.}

Polar coordinates explicitly encode the change in heading angle ($\varphi$). This intuitively encodes a \textit{smooth, non-holonomic} motion prior.
For example, polar features for trajectory $\overrightarrow{KL}$ and an (improbable) trajectory $\overrightarrow{MN}$ significantly differ, while their Cartesian coordinates do not. 
These key characteristics allow \methodabr to generalize well even when trained with a small number of labeled samples (Sec. \ref{sec:exp_generalization}), especially important for rarely observed object classes. 
%

\input{tables/tab_nuscenes_test_benchmark}

\input{tables/tab_nuscenes_val_benchmark}

\subsection{Online Graph Construction}
\label{subsec:online_tracking}

In this section, we propose an online graph construction approach that makes our method applicable to \textit{online} applications. 
We maintain a \textit{single} input graph for the whole sequence, which we continuously evolve with each incoming frame. This construction is identical to the \textit{offline} setting, as discussed in Sec. \ref{subsec:graph_scene_representation}. In each frame, edges are classified based on past-frame information \textit{only}. 
There are multiple ways of maintaining a single input graph over a sequence of frames. 

\PAR{MPNTrack++: dense.} First approach is a simple extension of prior offline GNN methods~\cite{Braso20CVPR,Li_2020_WACV}, where previous track associations do not influence the graph structure. At each frame, new nodes and edges are added to a continuously growing \textit{dense} graph. \textit{Such model is unaware of past track estimates.} 

\PAR{Prune inactive: consecutive.} Past track estimates are a valuable cue that could be leveraged to resolve ambiguous associations. To this end, \cite{zaech2022learnable} propose to simply prune all past (negative) edges. 
This option encodes track history and maintains sparsity and due to frame-by-frame processing, preserved track edges always connect temporally \textit{consecutive} nodes in each track. 
However, over time, the temporal distance from the current frame to early track nodes is increasing, making them unreachable within a limited number of message passing iterations. \textit{This approach thus has a limited temporal receptive field}.

\PAR{Ours: prune + skip.} We propose a simple solution, that retains graph sparsity and provides global temporal receptive field from each newly-added node. After each frame, we remove all \textit{past negative} edges \textit{and} ensure we have, for each track, an edge between \textit{each} most-recent track node and \textit{all} past nodes from its track, see Fig.~\ref{subfig:overview_online}. 
Moreover, new nodes (detections) at each frame are \textit{only} connected to the most-recent track node for each existing track. 
By continuously evolving the graph in this manner we maintain sparsity, provide previous tracking decisions to the model through the input graph topology (this makes our model autoregressive) and keep all nodes reachable during message passing: any two connected nodes have at most two edges between them. 
\textit{This allows our model to \textit{learn} historical context directly via message passing to make well-informed edge classification, as needed for long-term tracking.}
%


\subsection{Implementation details}
\label{subsec:implementation} 

\PAR{Network structure.} 
In \methodabr, we use only fully connected (FC) layers with the leaky ReLU \cite{Nair2010RectifiedLU} nonlinearity, and the dimensionality of the edge and node features is $16$ and $32$ respectively.
Our $\mlp_{node}$ and $\mlp_{node\_init}$ consist of 3 FC layers. Remaining MLPs consist of 2 layers with 70k parameters in total.

\PAR{Training and augmentation.} 
We train \methodabr only on keyframes from the nuScenes \cite{Caesar20CVPR} dataset, which we augment with noise dropout to mimic real detection performance. 
To mimic occlusions and false negatives, we randomly drop full frames, individual bounding boxes, and edges. 
To mimic false positives, we randomly add bounding boxes. 
We also perturb boxes by adding Gaussian noise to 3D positions and orientation. 
We train \methodabr with focal loss \cite{lin2017focal}, which is well suited for our imbalanced binary classification case. We use Radam optimizer~\cite{liu2019radam,Kingma15ICLR} using cosine annealing with warm restarts \cite{DBLP:conf/iclr/LoshchilovH17} for $180$ epochs with batch size of $64$. 
For more details, we refer to the supplementary. 

\section{Experimental Evaluation}
\label{sec:experimental}

This section outlines our evaluation setting: datasets, metrics and input detections (Sec.~\ref{sec:eval_set}). 
Next, in Sec.~\ref{sec:bench}, we discuss our \textit{offline} tracking results on the nuScenes test set and \textit{online} and \textit{offline} performance on the validation set, compared to the current state-of-the-art.
Then, we justify our design decisions and discuss the benefits of our contributions via thorough ablation studies (Sec.~\ref{sec:ablation}).
Finally, we demonstrate the generalization capabilities of \methodabr by successfully applying it to different datasets and locations without fine-tuning (Sec.~\ref{sec:exp_generalization}).

\subsection{Evaluation setting}
\label{sec:eval_set}

\input{tables/tab_nuscenes_offline_ablation_invariances_mini}
\PAR{Datasets.} 
We evaluate our method on nuScenes~\cite{Caesar20CVPR} and KITTI~\cite{Geiger12CVPR} tracking datasets. NuScenes was recorded in four locations across two cities: Boston, USA, and Singapore. It contains 150 scenes, recorded with a 32-beam lidar sensor, and provides two scans per second (2Hz). 
KITTI tracking dataset was recorded in Karlsruhe, Germany, using a 64-beam lidar sensor at a 10Hz frame rate.
Unless otherwise specified, we follow the official train/validation/test splits.

\PAR{Evaluation metric.}
On the nuScenes dataset, we follow the official evaluation protocol that reports per-class and average AMOTA~\cite{Weng20iros}, which averages the CLEAR-MOT~\cite{Bernardin08JIVP} MOTA metric across different recall thresholds. 
On KITTI, we also report sAMOTA~\cite{Weng20iros} and HOTA~\cite{luiten20ijcv} metrics, which allows us to analyze detection and association errors separately. 
We also report the number of identity switches (IDs) 
at the best performing recall. 
In the supplementary, we provide extended tables and evaluations that we omitted for brevity.

\PAR{Object detections.} 
On nuScenes, for a fair comparison with CenterPoint~\cite{yin2021center}, we use their provided 3D detections (without the estimated velocity vectors). On KITTI, we use PointGNN~\cite{shi2020point} detections provided by EagerMOT~\cite{Kim21ICRA}.

\subsection{Benchmark results}
\label{sec:bench}

We compare our method to state-of-the-art 3D MOT on the official nuScenes benchmark. In Tab.~\ref{tab:nuscenes_sota_test} (\textit{top}) we compare methods that rely \textit{only} on the 3D input. As can be seen, our model ranks highest overall ($66.4$ avg. AMOTA compared to 2nd best $65.6$). Our method only lags behind methods that additionally rely on rich visual signal (AlphaTrack and EagerMOT). While our focus was on effectively leveraging geometric relations, this hints that our approach could potentially further benefit from sensor fusion, which we leave for future work.

\PAR{Online tracking.}
To confirm that our approach successfully handles online scenarios, we evaluate it on the nuScenes validation set with streaming inputs. 
In Tab.~\ref{tab:nuscenes_sota_val} we compare our \textit{offline} and \textit{online} versions, and compare both to \textit{online} state-of-the-art CenterPoint~\cite{yin2021center} using the same 3D detections for all.

Not surprisingly, our \textit{offline} model achieves the top performance at $71.14$ avg. AMOTA. When switching to \textit{online} inference, we achieve $67.27$ avg. AMOTA ($-3.87$), outperforming CenterPoint by $+1.36$ avg AMOTA (from $65.91$).
As expected, there is a noticeable difference between our \textit{offline} and \textit{online} variants, as the online version is effectively exposed to only half of the context available to its offline counterpart, which observes both past and future detections.

\subsection{Model ablation}
\label{sec:ablation}

\PAR{Edge parametrization.} 
To measure the impact of our proposed representation of geometric relations, we train three models with different relative feature parametrizations. 
The main advantage of our \textit{proposed polar} representation is in the inductive bias that helps the model better understand long trajectories and non-holonomic motion. We conduct this experiment in the low data regime using the official nuScenes mini-split (1\% of training data) and evaluate them on the (non-overlapping) validation set (150 seq.) and report results in Tab.~\ref{tab:nuscenes_offline_invariances_mini}. 


First row reports results obtained with our proposed model (time-normalized localized polar parametrization, together with orientation and time difference, as explained in Sec.~\ref{sec:polar_encoding}). This approach yields $57.96$ avg. AMOTA.
Next is the model without time normalization, where the features include distance between nodes instead of velocity. Despite the time difference still being available in this configuration, we see that explicit time normalization yields an improvement of \textbf{$+5.79$} avg. AMOTA, confirming our intuition that encoding relative velocity instead of distance improves tracking by making features invariant to trajectory length.
Lastly, we evaluate a model with the commonly-used global Cartesian representation and see that simply switching to our proposed localized polar parametrization gives a significant improvement of \textit{$+17.55$} avg. AMOTA. This validates our intuition that incorporating a domain-appropriate inductive bias improves generalization from scarce data. 
%
%
For completeness, in the supplementary, we provide evaluations of these models trained on the full training set.

\PAR{Intra-frame connections.} 
\textit{Does spatial context matter?} Tab.~\ref{tab:nuscenes_offline_ablation_intraframe_edges} shows that adding spatial \textcolor{darkred}{\textit{intra-frame}} edges improves avg. AMOTA by $+1.05$
and significantly improves recall ($+2.4\%$) at a marginal increase in IDs ($+15$). 
For better contextual awareness, we aggregate messages from temporal and spatial edges separately. 
An ablation study of this technique is in the supplementary material. 

\input{tables/tab_nuscenes_offline_ablation_intraframe_edges}
\input{tables/tab_nuscenes_online_ablation_max_edge_distance}

\PAR{Sparse graph construction.} 
\textit{What is the impact of our sparse graph construction on tracking performance?} 
In Tab.~\ref{tab:nuscenes_online_ablation_max_edge_distance} we compare results obtained by models using half ($0.5$x) and double ($2.0$x) of the measured maximal velocity values to constrain edges with the ones from our default model ($1$x). 
Reducing maximum allowed velocity ($0.5$x) removes valuable edges that hypothesize valid associations, while permitting higher values ($2$x) introduces irrelevant edges. 
In both cases, we observe a significant performance drop across all metrics (recall, precision, AMOTA), due to message passing becoming less complete and noisier and edge classification more challenging.
This confirms that our data-driven physics-based sparse graph construction is the optimal approach.

\input{tables/tab_nuscenes_online_ablation_connectivity}
\PAR{Evolving online graph connectivity.}
To validate our proposed approach to online graph connectivity, we present ablation experiments that confirm the advantage of our technique over alternatives discussed in Sec.\ref{subsec:online_tracking}. 
As shown in Tab.~\ref{tab:nuscenes_online_ablation_connectivity}, our experimental results align with our expectations based on the theoretical properties of each option. 
\textit{Dense} (MPNTrack++) linking ($61.39$ avg. AMOTA) completely ignores past trajectories and produces the lowest results. 
Next, \textit{consecutive} (prune inactive) chaining introduces a significant improvement with $66.03$ avg. AMOTA.
Finally, our proposed \textit{prune + skip} connectivity leads to a significant improvement over both alternatives with $67.27$ avg. AMOTA.

\subsection{Generalization study}
\label{sec:exp_generalization}

In this section, we show how \methodabr generalizes across geographic locations and datasets. To this end, we (i) use the nuScenes dataset, recorded in Singapore and Boston, 
to train the model on one city and evaluate on the other
and (ii) evaluate our model, trained on nuScenes, on KITTI data without any fine-tuning.

\input{tables/tab_nuscenes_offline_generalization_cities}
\PAR{Cross-city generalization: Boston $\leftrightarrow$ Singapore.} In this cross-city evaluation we compare our method to CenterPoint (we use detections from the corresponding re-trained CenterPoint model). As shown in Tab.~\ref{tab:nuscenes_generalization_cities}, our method generalizes significantly better: 
$+3.41$ avg. AMOTA
on Boston (trained) $\to$ Singapore (evaluated) and 
$+3.33$ avg. AMOTA
on Singapore $\to$ Boston. 
In both cases we also observe a significant improvement in both recall and number of ID-switches. 
This confirms that our method generalizes very well across regions, which we believe is largely due to our feature parametrization (Sec.~\ref{sec:polar_encoding}).

\input{tables/tab_kitti_val_benchmark}
\input{tables/tab_kitti_test_benchmark}
\PAR{Cross-dataset generalization: nuScenes $\to$ KITTI.} 
We report our results on the unofficial KITTI 3D MOT benchmark~\cite{Weng20iros} in Tab.~\ref{tab:kitti_sota_val}, and on the official KITTI 2D MOT benchmark in Tab.~\ref{tab:kitti_sota_test}. 
We \textit{only} evaluate our nuScenes-trained model on KITTI dataset, which we consider an ultimate generalization experiment, as KITTI dataset was recorded using a different sensor under a different frame rate in a different geographical location (Karlsruhe, Germany). 
For 2D MOT evaluation we simply project our estimated 3D tracks to the image plane.
%
We note that entries are not directly comparable, as different methods use different input detections. However, we observe our method is on-par with EagerMOT~\cite{kim20cvpr}, which uses lidar and cameras, and the same set of 3D detections. \methodabr is top-performer on the 3D benchmark and among top-4 on the 2D benchmark, performing consistently well on \textit{car} and \textit{pedestrian} classes. 

%
\section{Conclusion}

We presented \methodabr for 3D multi-object tracking that solely relies on 3D bounding boxes as input without appearance/shape information. Our key contribution is a GNN that encodes spatial and temporal geometric relations via localized polar coordinates. This parametrization enables us to effectively learn to understand long-range temporal and spatial context via message passing and, solely from object interactions, learn a scene representation suitable for tracking via edge classification. 
We also propose an online graph construction technique to apply \methodabr to streaming data. 
Our method establishes a new state-of-the-art on the nuScenes dataset among methods that do not rely on image data and, more importantly, generalizes well across geographic regions and datasets. 

\PAR{Acknowledgments.}
This research was partially funded by the Humboldt Foundation through the Sofja Kovalevskaja Award. 

\clearpage
%
%
\bibliographystyle{splncs04}
\bibliography{abbrev_short,refs}

\clearpage
{
\centering
\Large
\textbf{PolarMOT: How far can geometric relations take
us in 3D multi-object tracking?} \\
\vspace{0.5em}Supplementary Material \\
\vspace{1.0em}
}
\appendix
\input{supplementary}

\end{document}

%% file: macros.tex
\newcommand*{\eg}{\emph{e.g.}\@\xspace}
\newcommand*{\ie}{\emph{i.e.}\@\xspace}

\makeatletter
\newcommand*{\etc}{%
    \@ifnextchar{.}%
        {etc}%
        {etc.\@\xspace}%
}

\newcommand{\PAR}[1]{\vskip4pt \noindent {\bf #1~}}

\definecolor{lightblue}{RGB}{46, 134, 193}
\definecolor{darkred}{RGB}{169, 50, 38 }

\newcommand*{\methodabr}{{\textit{PolarMOT}}\@\xspace}
 
\newcommand*{\papertitle}{{PolarMOT: How Far Can Geometric Relations Take Us in 3D Multi-Object Tracking?}\@\xspace}
\newcommand*{\githuburl}{polarmot.github.io}

\newcommand*{\relativefeaturescapital}{{{Localized relational polar encoding}}\@\xspace}

\newcommand*{\mlp}{{\text{MLP}}}

\newcommand{\gray}[1]{\textcolor{gray}{#1}}

\newcommand{\crossmark}{\ding{55}}

\def\unknown{\gray{unpub}\space}

\newcommand{\B}[1]{\textbf{#1}}
\newcommand{\I}[1]{\emph{#1}}

%% file: tables/tab_nuscenes_test_benchmark.tex
\begin{table}[t]
\centering
\setlength{\tabcolsep}{2.7pt}
\ssmall
%
\caption{Results of state-of-the-art methods for 3D multi-object tracking on the NuScenes test set. Legend: L -- lidar, B -- 3D bounding boxes}
\label{tab:nuscenes_sota_test}
    \begin{tabular}{l l| c c c | c c c c c c c}
    \toprule
    \multirow{2}{*}{Method name} & \multicolumn{1}{l|}{Input} & IDs $\downarrow$ & Recall $\uparrow$ & AMOTA $\uparrow$ & \multicolumn{7}{c}{class-specific AMOTA $\uparrow$} \\
     & \multicolumn{1}{c|}{modality} & total & average & average &
    car & ped & bicycle & bus & motor & trailer & truck  \\
    \midrule
    Ours 
    & 3D (B)
    & \B{242}
    & \B{70.2}
    & \B{66.4}
    & \B{85.3}
    & \B{80.6}
    & 34.9
    & {70.8} 
    & \B{65.6}
    & \B{67.3}
    & {60.2} 
    \\
    OGR3MOT \cite{zaech2022learnable}
    & 3D (B)
    & 288
    & 69.2
    & 65.6
    & 81.6
    & 78.7
    & \B{38.0}
    & \B{71.1}
    & 64.0
    & \B{67.1}
    & 59.0 
    \\
    CenterPoint \cite{yin2021center}
    & 3D (L)
    & 684
    & 68.0
    & 65.0
    & 81.8
    & 78.0
    & 33.1
    & {71.5}
    & 58.7
    & 69.3
    & \B{62.5}
    \\
    IPRL-TRI  \cite{chiu2020probabilistic}
    & 3D (B)
    & 950
    & 60.0
    & 55.0
    & 71.9
    & 74.5
    & {25.5}
    & {64.1}
    & 48.1
    & 49.5
    & 51.3 
    \\
    \midrule
    AlphaTrack\cite{zeng21iros}
    & 3D + 2D
    & 718
    & 72.3
    & {69.3}
    & {84.2}
    & {74.3}
    & 47.1
    & {72.0} 
    & 72.8
    & 72.0 
    & {62.6} 
    \\

    EagerMOT \cite{Kim21ICRA}
    & 3D + 2D
    & 1156
    & 72.7
    & {67.7}
    & 81.0
    & 74.4
    & {58.3}
    & 74.0
    & {62.5}
    & 63.6
    & 59.7 
    \\
    \bottomrule
    \end{tabular}
%
\end{table}

%% file: tables/tab_nuscenes_val_benchmark.tex
\begin{table}[t]
\centering
\ssmall
\setlength{\tabcolsep}{2.0pt}
\caption{Online vs. offline tracking on the nuScenes validation set~\cite{Caesar20CVPR}}
\label{tab:nuscenes_sota_val}
    \begin{tabular}{l l| c c c| c c c c c c c}
    \toprule
    \multirow{2}{*}{Method name} & \multicolumn{1}{l|}{Input} & IDs $\downarrow$ & Recall $\uparrow$ & AMOTA $\uparrow$ & \multicolumn{7}{c}{class-specific AMOTA $\uparrow$} \\
     & \multicolumn{1}{c|}{modality} & total & average & average &
    car & ped & bicycle & bus & motor & trailer & truck  \\
    \midrule
    %
    %
    Ours \I{offl.} 
    & 3D
    & \B{213}
    & 75.14
    & \B{71.14}
    & \B{85.83}
    & \B{81.70}
    & 54.10 
    & \B{87.36} 
    & 72.32
    & 48.67 
    & \B{68.03} 
    \\
%
    Ours \I{onl.} 
    & 3D
    & 439 
    & 72.46
    & 67.27
    & 81.26 
    & 78.79
    & 49.38
    & 82.76
    & 67.19
    & 45.80 
    & 65.70 
    \\
    CenterPoint \I{onl.}
    & 3D
    & 562
    & 70.62
    & 65.91
    & 84.23
    & 77.29
    & 43.70
    & 80.16
    & 59.16
    & \B{51.47}
    & 65.39 
    \\
    \bottomrule
    \end{tabular}
\end{table}
%

%% file: tables/tab_nuscenes_offline_ablation_invariances_mini.tex
\begin{table*}[t]
\centering
\setlength{\tabcolsep}{1.7pt}
\scriptsize
\caption{Ablation on parametrization of geometric relations among objects on nuScenes validation set. Trained on the \textbf{official mini} training set}
\label{tab:nuscenes_offline_invariances_mini}
    \begin{tabular}{c c | c c c | c c c c c c c}
    \toprule
    %
    %
    %
    %
    Localized & Normalized
    & IDs $\downarrow$ & Recall $\uparrow$ & AMOTA $\uparrow$ & \multicolumn{7}{c}{class-specific AMOTA $\uparrow$} \\
    polar & by time
    & total & average & average &
    car & ped & bicycle & bus & motor & trailer & truck \\
    \midrule
    %
    \checkmark 
    & \checkmark
    & 430
    & \B{62.12}
    & \B{57.96}
    & \B{85.16}
    & \B{80.80}
    & \B{44.02}
    & \B{80.68}
    & 45.83
    & \B{5.41}
    & \B{63.83}
    \\
    \checkmark 
    & \textcolor{red}{\crossmark}
    & 652
    & 55.07
    & 52.17
    & 81.51
    & 80.14
    & 06.66
    & 79.53
    & 53.82
    & 0
    & 63.51 
    \\
    \textcolor{red}{\crossmark} 
    & \checkmark
    & 1321
    & 41.06
    & 40.41
    & 78.77
    & 78.80
    & 0
    & 67.62 
    & 0
    & 0
    & 57.66 
    \\
    \bottomrule
    \end{tabular}
\end{table*}

%% file: tables/tab_nuscenes_offline_ablation_intraframe_edges.tex
\begin{table}[t]
\centering
\setlength{\tabcolsep}{2.5pt}
\scriptsize
\caption{Ablation for intra-frame edges on the nuScenes validation set}
\label{tab:nuscenes_offline_ablation_intraframe_edges}
    \begin{tabular}{c | c c c | c c c c c c c}
    \toprule
    {Intra-frame} & IDs $\downarrow$ & Recall $\uparrow$ & AMOTA $\uparrow$ & \multicolumn{7}{c}{class-specific AMOTA $\uparrow$} \\
    connections & total & average & average &
    car & ped & bicycle & bus & motorcycle & trailer & truck  \\
    \midrule
    \checkmark 
    & 213
    & \B{75.14}
    & \B{71.14}
    & \B{85.83}
    & \B{81.70}
    & \B{54.10}
    & \B{87.36}
    & \B{72.32}
    & \B{48.67}
    & 68.03 
    \\
    \textcolor{red}{\crossmark} 
    & \B{198}
    & {72.74}
    & 70.09
    & 85.44
    & 80.51
    & 52.88
    & 86.78
    & 69.87
    & 46.61
    & \B{68.54} 
    \\
    \bottomrule
    \end{tabular}
\end{table}

%% file: tables/tab_nuscenes_online_ablation_max_edge_distance.tex
\begin{table}[t]
\centering
\setlength{\tabcolsep}{3.0pt}
\tiny
\caption{Sparse graph construction: the impact of reducing/increasing the maximal velocity threshold on online tracking (nuScenes validation set)}
\label{tab:nuscenes_online_ablation_max_edge_distance}
    \begin{tabular}{c | c c c c | c c c c c c c}
    \toprule
  %
%
    \multicolumn{1}{c|}{Max edge} & IDs $\downarrow$ & Recall $\uparrow$ & AMOTP $\downarrow$ & AMOTA $\uparrow$ & \multicolumn{7}{c}{class-specific AMOTA $\uparrow$} \\
    distance & total & average & average &
    car & ped & bicycle & bus & motorcycle & trailer & truck  \\
    \midrule
%
    0.5x
    & {1123} 
    & 65.14
    & 0.718
    & {58.61} 
    & {76.90} 
    & {46.66} 
    & {46.89} 
    & {77.35} 
    & {62.73} 
    & {40.09} 
    & {59.63} 
    \\
  %
    \B{1.0x}
    & \B{439}
    & \B{72.46}
    & \B{0.595}
    & \B{67.27}
    & \B{81.26}
    & \B{78.79}
    & \B{49.38}
    & {82.76}
    & \B{67.19}
    & \B{45.80}
    & \B{65.70}
    \\
%
    2.0x
    & 467
    & 69.74
    & 0.642
    & {65.42}
    & \B{81.28}
    & {72.15}
    & {46.84}
    & \B{83.45}
    & {63.57}
    & {44.94}
    & \B{65.70}
    \\
%
%
    \bottomrule
    \end{tabular}
%
\end{table}

%% file: tables/tab_nuscenes_online_ablation_connectivity.tex
\begin{table}[t]
    \centering
    \scriptsize
    \caption{Online graph construction analysis (nuScenes validation set)}
    \label{tab:nuscenes_online_ablation_connectivity}
        \begin{tabular}{c | c c c | c c c c c c c}
        \toprule
    %
    %
    %
        Track & IDs $\downarrow$ & Recall $\uparrow$ & AMOTA $\uparrow$ & \multicolumn{7}{c}{class-specific AMOTA $\uparrow$} \\
        connectivity & total & average & average &
        car & pedestrian & bicycle & bus & motorcycle & trailer & truck  \\
    %
        \midrule 
    %
        \B{Ours}
        & \B{439}
        & \B{72.46}
        & \B{67.27}
        & \B{81.26}
        & \B{78.79}
        & 49.38
        & \B{82.76}
        & \B{67.19}
        & \B{45.80}
        & \B{65.70}
        \\
        Consecutive
        & 485
        & 69.90
        & 66.03
        & 81.04
        & 77.37
        & 48.84
        & 82.45
        & 66.58
        & 40.66
        & 65.23
        \\
        Dense
        & 1024
        & 68.58
        & 61.39
        & 71.08
        & 75.20
        & \B{49.70}
        & 79.78
        & 52.03
        & 41.15
        & 60.79
        \\
        \bottomrule
        \end{tabular}
    \end{table}

%% file: tables/tab_nuscenes_offline_generalization_cities.tex
\begin{table}[t]
\centering
\ssmall
\setlength{\tabcolsep}{2.4pt}
\caption{
    CenterPoint (CP) \cite{yin2021center} and our method when trained on training data from one city, and evaluated on the validation data from another}
\label{tab:nuscenes_generalization_cities}
    \begin{tabular}{l | c | c c c| c c c c c c c}
    \toprule
%
    \multicolumn{1}{l|}{Train city} & \multicolumn{1}{c|}{Tracking} & IDs $\downarrow$ & Recall $\uparrow$ & AMOTA $\uparrow$ & \multicolumn{7}{c}{class-specific AMOTA $\uparrow$} \\
    \multicolumn{1}{l|}{$\to$ eval city} & \multicolumn{1}{c|}{model} & total & average & average &
    car & ped & bicycle & bus & motor & trailer & truck  \\
    \midrule
    %
    %
    Boston
    & Ours 
    & \B{145}
    & \B{64.48}
    & \B{63.12}
    & \B{82.26}
    & \B{72.81}
    & \B{31.49} 
    & {77.70} 
    & \B{43.17}
    & 0 
    & \B{71.28} 
    \\
%
    %
%
    $\to$ Singapore
    & CP
    & 306
    & 61.02
    & 59.71
    & 79.26
    & 67.47
    & 20.52
    & \B{78.86}
    & 41.13
    & 0
    & 71.04 
    \\
    \midrule
%
    %
    Singapore
    & Ours 
    & \B{104}
    & 52.30
    & \B{50.28}
    & \B{78.60}
    & \B{82.59}
    & \B{36.70} 
    & \B{71.22} 
    & \B{28.71}
    & 11.31
    & {42.83} 
    \\
    %
    %
%
    $\to$ Boston
    & CP
    & 314
    & \B{53.32}
    & 47.06
    & 77.01
    & 76.18
    & 34.86
    & 71.07
    & 13.54
    & \B{13.18}
    & \B{43.55} 
    \\
    \bottomrule
    \end{tabular}
\end{table}
%

%% file: tables/tab_kitti_val_benchmark.tex
\begin{table}[t]
\centering
\setlength{\tabcolsep}{4.6pt}
\ssmall
\caption{Unofficial KITTI 3D MOT validation set benchmark~\cite{Weng20iros}. Our model was trained \textbf{only} on nuScenes dataset}
\label{tab:kitti_sota_val}
%
    \begin{tabular}{l c c | c c | c c | c c | c c}
    \toprule
    \multirow{2}{*}{Method name} & {3D} & 2D & \multicolumn{2}{c|}{IDs $\downarrow$} & \multicolumn{2}{c|}{sAMOTA $\uparrow$} & \multicolumn{2}{c|}{MOTA $\uparrow$} & \multicolumn{2}{c}{Recall $\uparrow$} \\
     & {input} & {input} & car & ped & car & ped & car & ped & car & ped \\
    \midrule
 %
    Ours \textit{online}
    & \checkmark
    & \textcolor{red}{\crossmark} 
    & {31}
    & {9}
    & {94.32}
    & \B{94.08}
    & {93.93} 
    & \B{93.48}
    & {94.54}
    & \B{93.66} 
    \\
    PC-TCNN \cite{wu2021}
    & \checkmark
    & \textcolor{red}{\crossmark} 
    & \B{1}
    & {--}
    & \B{95.44}
    & {--}
    & \unknown
    & {--}
    & \unknown
    & {--} 
    \\
    EagerMOT \cite{Kim21ICRA}
    & \checkmark
    & \checkmark
    & \B{2}
    & {36}
    & {94.94}
    & {92.95}
    & \B{96.61} 
    & {93.14}
    & \B{96.92}
    & \B{93.61} 
    \\
    GNN3DMOT \cite{weng20CVPR}
    & \checkmark 
    & \checkmark
    & {10}
    & {--}
    & {93.68}
    & {--}
    & {84.70} 
    & {--}
    & {\unknown}
    & {--} 
    \\
    AB3DMOT \cite{Weng20iros} 
    & \checkmark
    & \textcolor{red}{\crossmark} 
    & \B{0}
    & \B{1}
    & {91.78}
    & {73.18}
    & {83.35} 
    & {66.98}
    & {92.17}
    & {72.82} 
    \\
    \bottomrule
    \end{tabular}
%
\end{table}

%% file: tables/tab_kitti_test_benchmark.tex
\begin{table}[t]
\centering
\setlength{\tabcolsep}{3.0pt}
\ssmall
\caption{KITTI 2D MOT test set benchmark. Our model was trained \textbf{only} on nuScenes dataset}
\label{tab:kitti_sota_test}
    \begin{tabular}{l c c | c c | c c | c c | c c | c c}
    \toprule
    \multirow{2}{*}{Method name} & {3D} & 2D & \multicolumn{2}{c|}{IDs $\downarrow$} & \multicolumn{2}{c|}{HOTA $\uparrow$} & \multicolumn{2}{c|}{AssA $\uparrow$} & \multicolumn{2}{c|}{AssRe $\uparrow$} & \multicolumn{2}{c}{MOTA $\uparrow$} \\
     & {input} & {input} & car & ped & car & ped & car & ped & car & ped & car & ped \\
    \midrule
    Ours \textit{online}
    & \checkmark
    & \textcolor{red}{\crossmark} 
    & {462}
    & {270}
    & {75.16}
    & {43.59}
    & {76.95} 
    & {48.12}
    & {80.00}
    & {51.95} 
    & {85.08}
    & {46.98} 
    \\
    PC-TCNN \cite{wu2021}
    & \checkmark
    & \textcolor{red}{\crossmark} 
    & \B{37}
    & {--}
    & \B{80.90}
    & {--}
    & \B{84.13} 
    & {--}
    & \B{87.46}
    & {--} 
    & \B{91.70}
    & {--} 
    \\
    PermaTrack \cite{tokmakov2021learning}
    & \textcolor{red}{\crossmark} 
    & \checkmark
    & {258}
    & {403}
    & {78.03}
    & {48.63}
    & {78.41} 
    & {45.61}
    & {81.14}
    & {49.63} 
    & {91.33}
    & {65.98} 
    \\
    PC3T \cite{wu20213d}
    & \checkmark
    & \textcolor{red}{\crossmark} 
    & {225}
    & --
    & {77.80}
    & --
    & {81.59} 
    & --
    & {84.77}
    & -- 
5
    & {88.81}
    & -- 
    \\
    Mono\_3D\_KF \cite{9626850}
    & \textcolor{red}{\crossmark} 
    & \checkmark
    & {162}
    & {267}
    & {75.47}
    & {42.87}
    & {77.63} 
    & {46.31}
    & {80.23}
    & {52.86} 
    & {88.48}
    & {45.44} 
    \\
    EagerMOT \cite{Kim21ICRA}
    & \checkmark
    & \checkmark
    & {239}
    & {496}
    & {74.39}
    & {39.38}
    & {74.16} 
    & {38.72}
    & {76.24}
    & {40.98} 
    & {87.82}
    & {49.82} 
    \\
    {SRK\_ODESA} \cite{ODESA2020}
    & \textcolor{red}{\crossmark} 
    & \checkmark
    & {380}
    & {511}
    & {68.51}
    & \B{50.87}
    & {63.08} 
    & {48.78}
    & {65.89}
    & {53.45} 
    & {87.79}
    & \B{68.04} 
    \\
    3D-TLSR \cite{nguyen20203d}
    & \textcolor{red}{\crossmark} 
    & \checkmark
    & {--}
    & \B{175}
    & {--}
    & {46.34}
    & {--} 
    & \B{51.32}
    & {--}
    & \B{54.45} 
    & {--}
    & {53.58} 
    \\
    MPNTrack \cite{Braso20CVPR}
    & \textcolor{red}{\crossmark} 
    & \checkmark
    & {--}
    & {397}
    & {--}
    & {45.26}
    & {--} 
    & {47.28}
    & {--}
    & {52.18} 
    & {--}
    & {46.23} 
    \\
    \bottomrule
    \end{tabular}
%
\end{table}

%% file: supplementary.tex
\section{Implementation details}

\subsection{Training and augmentation}

\PAR{Training.}
As training data we use annotated training keyframes from the nuScenes~\cite{Caesar20CVPR} dataset. We represent labeled boxes as nodes in the graph. Edges in the graph are labeled positive if they connect nodes with the same track ID (across any number of frames) and negative otherwise. 
During model training, all input clips are processed individually and each edge is considered an independent sample that contributes to the total focal loss~\cite{lin2017focal}.



\PAR{Data augmentation.}
To mimic noisy, real-world detectors, we rely on data augmentation. 
We add random bounding box detections at each frame before the graph construction to imitate false positive detections. 
For each frame, the number of added boxes is a fraction of the number of real boxes (between 0.7 and 0.9 for each class) plus a fixed number (between 1 and 3) to augment completely empty frames. 
Pose coordinates (position and orientation) of augmented boxes are sampled from uniform distributions whose parameters are the minimum and maximum values of corresponding coordinates to labeled boxes.

We also augment input graphs at each training iteration. 
To mimic occlusions and miss-classifications, we randomly drop nodes (between 40\%-60\%) at each frame as well as some complete frames from our graphs.
To emulate imprecise detections, we further perturb each initial edge feature, representing differences in object poses, with noise vectors sampled from class-specific Gaussian distributions with zero mean.
For each class, the standard deviation for distributions of distance noise (meters) is between $0.05$ and $0.35$, for polar angle noise (radians) between 0.1 and 0.25, and for orientation noise (radians) between 0.05 and 0.25. 
For further augmentation, we fully remove approximately 20\% of all edges in the graph. 
These augmentations ensure that our model is robust to imperfect/noisy inputs that we obtain from real-world 3D object detectors. 

\PAR{Inference.}
In multi-object tracking, track IDs need to be distinct at each frame, \ie, only one object detection can be assigned to each identity (track ID). 
In our problem setting, this means that every node in the graph can have at most one positive edge connecting it to each past and future frame. We do not impose this constraint on \methodabr during the training (\ie, we train our model as an unconstrained binary classifier). However, we do ensure this constraint during inference via a simple post-processing procedure. %
%

%
In particular, given edge classification scores from our model, edges with a score higher than a certain threshold (between 0.5 and 0.8 for each class) are positive, others are negative.
Then, positive edge labels are \textit{greedily} assigned starting from the highest score. 
As soon as a positive edge is assigned between nodes $o_i^k$ and $o_j^m$ at frames $k$ and $m$, all other edges between node $o_i^k$ and frame $m$ (and between node $o_j^m$ and frame $k$) are ignored from further assignment.
This greedy procedure ensures that positive edges with the highest confidence are assigned first and each node has at most one positive edge to each frame.

\subsection{Network structure}

\input{tables/tab_architecture}

In this section, we detail the architecture of our network, explained in Sec. 4.2 in the main paper. 
In Tab.~\ref{tab:model_architecture}, we outline our network architecture composed entirely of multi-layer perceptrons (MLPs) with fully-connected layers. For each MLP, we list the dimensionality of all of its layers: input, intermediary and output.
Here, ``Edge initial'' and ``Node initial'' columns correspond to $\mlp_{edge\_init}$ and $\mlp_{node\_init}$ (Eq. 4 and 5 in the paper), which produce initial learned embeddings from the initial relative features. 
The ``Edge model'' column describes $\mlp_{edge}$ (Eq. 1 in the paper) that processes edge features at each message passing step. 
The columns ``Edge pres., past\&fut.'' describe the identical composition of \textcolor{darkred}{$\mlp_{pres}$} and \textcolor{lightblue}{$\mlp_{past}$/$\mlp_{fut}$}, which process \textcolor{darkred}{intra-frame} edges and two temporal directions of \textcolor{lightblue}{inter-frame} edges (Eq. 2 in the paper).
The ``Node model'' column outlines $\mlp_{node}$ that aggregates all edge embeddings and produces node features at each step (Eq. 3 in the paper).
Finally, the last column denotes the structure of the MLP used to classify edges based on the latest edge embeddings. We use leaky ReLU \cite{Maas2013RectifierNI} between all layers of the network. 

In our experiments, we always perform $L=4$ message passing steps. For offline inference, we process clips of 11 frames and for online tracking, we keep only the 3 latest detections for each track history. 



\section{Experimental evaluation}
 
\subsection{Edge parametrization ablation on the full training set}
\input{tables/tab_nuscenes_offline_ablation_invariances_full}
In the main paper, we ablated the impact of our proposed feature representation (time-normalized localized polar coordinates) by comparing models trained on the official nuScenes mini split. 
The main advantage of our representation is the inductive bias, which helps the model better understand long trajectories, turns and non-holonomic motion in general. The benefits of this inductive bias are best demonstrated when the amount of training data is low, which is why we used the mini split in our ablation.

For completeness, in Tab. \ref{tab:nuscenes_offline_invariances_full}, we provide the same ablation when the full training set is used. Unsurprisingly, with enough data (\eg ~{car, pedestrian and bus} classes), different feature representations perform similarly because there are enough samples to learn motion bias directly from data.
On the other hand, for rarely-observed classes, such as bicycles, motorcycles and trailers, using a better feature representation is clearly beneficial, \eg \textit{bicycle} AMOTA rises from $48.92$ to $54.10$.
This aligns with our main ablation results, where our parametrization outperforms standard representation in low data regimes.



\subsection{Contextual node aggregation}
\input{tables/tab_nuscenes_offline_ablation_aggregation_groups}
During message passing, our node aggregation step processes messages from \textcolor{lightblue}{past}, \textcolor{darkred}{present} and \textcolor{lightblue}{future} separately to maintain contextual awareness. To ablate the importance of this technique, we evaluate 3 versions of our trained model with different aggregation logic and show the results in Tab. \ref{tab:nuscenes_offline_ablation_aggregation_groups}. 

When both \textcolor{lightblue}{temporal} messages (past and future) are aggregated together, the model loses its time-awareness and its tracking performance significantly declines $-15.31$ avg. AMOTA (from $71.14$ down to $55.83$). 
Moreover, if \textcolor{darkred}{spatial} messages are also aggregated in the same single group, the model is completely unaware of scene context and geometric scene composition, so its avg. AMOTA falls further by $-32.53$ (a total decline of $47.84$ from our default contextual aggregation).

\subsection{Cross-city generalization oracle results}
\input{tables/tab_nuscenes_offline_generalization_cities_oracle}

In the main paper, we demonstrated the ability of our model to generalize across different cities by training it in Boston and evaluating in Singapore (and vice versa). 
To provide a better baseline and show how well the model would normally perform in each of the cities, we present Tab. \ref{tab:nuscenes_generalization_cities_oracle} where models are trained and evaluated on the same single city, \ie only Boston or only Singapore.

Since \methodabr demonstrates better performance than CenterPoint~\cite{yin2021center} on the full nuScenes~\cite{Caesar20CVPR} validation set (see Tab. 2 in the main paper), it is unsurprising that its results on individual cities are also better.

\subsection{Extended evaluation results}

\input{tables/tab_nuscenes_test_benchmark_ext}

\input{tables/tab_nuscenes_val_benchmark_ext}

\input{tables/tab_nuscenes_offline_generalization_cities_ext}

In this section, we present extended versions of the experimental evaluations detailed in the main paper.
These tables include additional tracking metrics reported by the nuScenes benchmark \cite{Caesar20CVPR}, which we provide for completeness:
\begin{itemize}
    \item MT (\textit{mostly tracked}): percentage of tracks tracked correctly for at least $80\%$ of their life span
    \item ML (\textit{mostly lost}) percentage of tracks tracked correctly for at most $20\%$ of their life span
    \item Frag. (\textit{fragmentations}): the number of times a trajectory is interrupted during tracking.
    \item TID (\textit{track initialization in seconds}): time until the first detection of the track is successfully tracked. 
    \item LGD (\textit{longest gap duration in seconds}): time an object instance has been incorrectly tracked.
\end{itemize}

In Tab.~\ref{tab:nuscenes_sota_test_ext}  we show extended results on the nuScenes test set benchmark (Tab. 9 in the main paper).

Tab.~\ref{tab:nuscenes_sota_val_ext} extends Tab. 2 in the main paper and details offline and online model evaluations on the nuScenes validation set. 

Tab.~\ref{tab:nuscenes_generalization_cities_ext} extends Tab. 7 in the main paper where \methodabr and CenterPoint \cite{yin2021center} are trained on one city and evaluated on the other. For evaluations of our method, we use detections produced by the corresponding CP model to make sure each pair of trackers uses the same set of detections.

\noindent \textbf{These extended evaluations along with our code, models and experimental data are available at \url{\githuburl}}.



%% file: tables/tab_architecture.tex
\begin{table}[t]
\centering
\setlength{\tabcolsep}{3.0pt}
\scriptsize
\caption{Neural network architecture of \methodabr. Each cell describes the output dimensionality of each layer in the fully-connected MLPs of our model}
\label{tab:model_architecture}
    \begin{tabular}{l | c c | c | c c | c}
    \toprule
    \multirow{2}{*}{} & \multicolumn{1}{c}{Edge} & Node & Edge & Edge & \multirow{1}{*}{Node} & Final edge \\

    {} & initial & initial & model & pres, past, fut & model & classifier \\
    \midrule
    Input
    & 4
    & 48
    & 80
    & 80
    & {96} 
    & {16}
    \\

    
    1st layer output
    & 16
    & 64
    & 64
    & {64}
    & 128 
    & {64}
    \\

    
    2nd layer output
    & 16
    & 128
    & {16}
    & {32}
    & {64} 
    & {32}
    \\

    
    3rd layer output
    & {}
    & {32}
    & {}
    & {}
    & {32} 
    & 16
    \\

    
    4th layer output
    & 
    & 
    & 
    & 
    & 
    & 1
    \\

    \bottomrule
    \end{tabular}
%
\end{table}

%% file: tables/tab_nuscenes_offline_ablation_invariances_full.tex
\begin{table*}[t]
\centering
\setlength{\tabcolsep}{1.7pt}
\scriptsize
\caption{Ablation on parametrization of geometric relations among objects on the nuScenes validation set. Trained on the \textbf{full} training set}
\label{tab:nuscenes_offline_invariances_full}
    \begin{tabular}{c c | c c c | c c c c c c c}
    \toprule
    Localized & Normalized
    & IDs $\downarrow$ & Recall $\uparrow$ & AMOTA $\uparrow$ & \multicolumn{7}{c}{class-specific AMOTA $\uparrow$} \\
    polar & by time
    & total & average & average &
    car & ped & bicycle & bus & motor & trailer & truck \\
    \midrule
    \checkmark 
    & \checkmark
    & 213
    & \B{75.14}
    & \B{71.14}

    & {85.83}
    & \B{81.70}
    & \B{54.10}
    & {87.36}
    & \B{72.32}
    & \B{48.67}
    & \B{68.03}
    \\
    %
    \checkmark 
    & \textcolor{red}{\crossmark}
    & \B{182}
    & 72.85
    & 70.27
    
    & \B{86.12}
    & \B{81.70}
    & 51.73
    & \B{87.79}
    & 69.29
    & 47.20
    & \B{68.07}
    \\
    \textcolor{red}{\crossmark} 
    & \checkmark
    & 225
    & 70.90
    & 69.75
    
    & 85.89
    & \B{81.72}
    & 48.92
    & 87.54
    & 69.25
    & 47.60
    & 67.31 
    \\
    \bottomrule
    \end{tabular}
\end{table*}

%% file: tables/tab_nuscenes_offline_ablation_aggregation_groups.tex
\begin{table}[t]
\centering
\setlength{\tabcolsep}{1.5pt}
\scriptsize
\caption{Ablation on the impact of contextual aggregation in node updates on the nuScenes validation set}
\label{tab:nuscenes_offline_ablation_aggregation_groups}
    \begin{tabular}{c | c c c | c c c c c c c}
    \toprule
    {Node aggregation} & IDs $\downarrow$ & Recall $\uparrow$ & AMOTA $\uparrow$ & \multicolumn{7}{c}{class-specific AMOTA $\uparrow$} \\
    connections & total & average & average &
    car & ped & bicycle & bus & motor & trailer & truck  \\
    \midrule
    Past/Present/Future
    & \B{213}
    & \B{75.14}
    & \B{71.14}
    & \B{85.83}
    & \B{81.70}
    & \B{54.10}
    & \B{87.36}
    & \B{72.32}
    & \B{48.67}
    & \B{68.03} 
    \\
  %
    Spatial/Temporal
    & 968
    & 60.29
    & 55.83
    & 81.48
    & 77.46
    & 48.76
    & 61.09
    & 30.23
    & 29.22
    & 62.56
    \\
  %
    All together 
    & 765
    & 26.75
    & 23.30
    & 0
    & 72.56
    & 40.75
    & 19.80
    & 0
    & 0
    & 30.04
    \\
    \bottomrule
    \end{tabular}
\end{table}

%% file: tables/tab_nuscenes_offline_generalization_cities_oracle.tex
\begin{table}[t]
\centering
\ssmall
\setlength{\tabcolsep}{2.4pt}
\caption{CenterPoint (CP) \cite{yin2021center} and our method when trained and evaluating only on one city}
\label{tab:nuscenes_generalization_cities_oracle}
    \begin{tabular}{l | c | c c c| c c c c c c c}
    \toprule
%
    \multicolumn{1}{l|}{Train city} & \multicolumn{1}{c|}{Tracking} & IDs $\downarrow$ & Recall $\uparrow$ & AMOTA $\uparrow$ & \multicolumn{7}{c}{class-specific AMOTA $\uparrow$} \\
    \multicolumn{1}{l|}{$\to$ eval city} & \multicolumn{1}{c|}{model} & total & average & average &
    car & ped & bicycle & bus & motor & trailer & truck  \\
    \midrule

    Singapore
    & Ours 
    & \B{145}
    & \B{77.82}
    & \B{75.79}
    & \B{85.42}
    & \B{77.17}
    & \B{51.41}
    & \B{85.76}
    & \B{74.23}
    & 0
    & \B{80.73}
    \\
    %
    %
%
    $\to$ Singapore
    & CP
    & 351
    & 72.97
    & 70.22
    & 82.66
    & 71.97
    & 39.71
    & 85.17
    & 62.36
    & 0
    & 79.44
    \\
    
    \midrule
    
    Boston
    & Ours 
    & \B{48}
    & \B{70.94}
    & \B{70.17}
    & \B{85.80}
    & \B{85.16}
    & \B{58.57}
    & \B{83.45}
    & \B{60.89}
    & \B{52.75}
    & \B{64.58}
    \\
%
%
    $\to$ Boston
    & CP
    & 246
    & 67.78
    & 65.96
    & 83.54
    & 81.37
    & 50.00
    & 82.47
    & 51.63
    & 48.61
    & 64.10
    \\
    \bottomrule
    \end{tabular}
\end{table}
%

%% file: tables/tab_nuscenes_test_benchmark_ext.tex
\begin{table}[t]
\centering
\setlength{\tabcolsep}{5.4pt}
\scriptsize
%
\caption{Extended results of state-of-the-art methods for 3D multi-object tracking on the NuScenes test set benchmark. Legend: L -- lidar, P -- ego poses, B -- 3D boxes}
\label{tab:nuscenes_sota_test_ext}
    \begin{tabular}{l l| c c c c c}
    \toprule
    \multirow{2}{*}{Method name} & \multicolumn{1}{c|}{Input} & MT $\uparrow$ & ML $\downarrow$ & Frag $\downarrow$ & TID $\downarrow$ & {LGD $\downarrow$} \\
     & \multicolumn{1}{c|}{modality} & total & total & total & average & average  \\
    \midrule
    Ours 
    & 3D (B)
    & \B{5701}
    & \B{1686}
    & \B{332}
    & {0.444}
    & \B{0.657}
    \\
    OGR3MOT \cite{zaech2022learnable}
    & 3D (B + P)
    & 5278
    & 2094
    & 371
    & 0.575
    & 0.782
    \\
    CenterPoint \cite{yin2021center}
    & 3D (L + P)
    & 5399
    & 1818
    & 553
    & \B{0.415}
    & 0.720
    \\
    IPRL-TRI  \cite{chiu2020probabilistic}
    & 3D (B + P)
    & 4294
    & 2184
    & 776
    & 0.960
    & 1.376
    \\
    
    \midrule
    
    AlphaTrack\cite{zaech2022learnable}
    & 3D + 2D + P
    & 5560
    & 1744
    & {480}
    & {0.409}
    & {0.755}
    \\
    EagerMOT \cite{Kim21ICRA}
    & 3D + 2D + P
    & 5303
    & 1842
    & {601}
    & 0.448
    & 0.801
    \\
    \bottomrule
    \end{tabular}
%
\end{table}

%% file: tables/tab_nuscenes_val_benchmark_ext.tex
\begin{table}[t]
\centering
\scriptsize
\setlength{\tabcolsep}{2.4pt}
\caption{Extended results for online vs. offline tracking on the nuScenes val set~\cite{Caesar20CVPR}}
\label{tab:nuscenes_sota_val_ext}
    \begin{tabular}{l l| c c c| c c c c c c c}
    \toprule
    \multirow{2}{*}{Method name} & \multicolumn{1}{l|}{Input} & MT $\uparrow$ & ML $\uparrow$ & Frag $\downarrow$ & TID $\downarrow$ & LGD $\downarrow$ \\
     & \multicolumn{1}{c|}{modality} & total & total & total &
    average & average \\
    \midrule

    Ours \I{offl.} 
    & 3D
    & \B{4524}
    & \B{1452}
    & {332}
    & \B{0.379}
    & \B{0.672}
    \\
%
    Ours \I{onl.} 
    & 3D
    & 4262
    & 1545
    & \B{285}
    & 0.636 
    & 0.901
    \\
    CenterPoint \I{onl.}
    & 3D
    & 4405
    & 1508
    & 445
    & 0.516
    & 0.956
    \\
    \bottomrule
    \end{tabular}
\end{table}
%

%% file: tables/tab_nuscenes_offline_generalization_cities_ext.tex
\begin{table}[t]
\centering
\scriptsize
\setlength{\tabcolsep}{2.4pt}
\caption{
    Extended results for CenterPoint (CP) \cite{yin2021center} and our method when trained on training data from one city, and evaluated on the validation data from another}
\label{tab:nuscenes_generalization_cities_ext}
    \begin{tabular}{l | c | c c c| c c c c c c c}
    \toprule
    \multicolumn{1}{l|}{Train city} & \multicolumn{1}{c|}{Tracking} & MT $\uparrow$ & ML $\downarrow$ & Frag $\downarrow$ & TID $\downarrow$ & LGD $\downarrow$ \\
    \multicolumn{1}{l|}{$\to$ eval city} & \multicolumn{1}{c|}{model} & total & total & total & average & average  \\
    \midrule
    Boston
    & Ours 
    & \B{1595}
    & \B{765}
    & \B{139}
    & \B{0.461}
    & \B{0.773}
    \\
    $\to$ Singapore
    & CP
    & 1464
    & 861
    & 224
    & 0.584
    & 1.037
    \\
    \midrule
    Singapore
    & Ours 
    & {2274}
    & 1177
    & \B{147}
    & {0.850}
    & \B{1.171}
    \\

    $\to$ Boston
    & CP
    & \B{2282}
    & \B{1147}
    & 253
    & \B{0.715}
    & 1.209
    \\
    \bottomrule
    \end{tabular}
\end{table}